\def\year{2018}\relax
\documentclass[letterpaper]{article} 
\usepackage{aaai18}  
\usepackage{times}  
\usepackage{helvet}  
\usepackage{courier}  
\usepackage{url}  
\usepackage{graphicx}  
\begin{document}
%
\title{Regularized Ensembles and Transferability in Adversarial Learning}
\author{Yifan Chen \\
Vanderbilt University\\
Nashville, TN 37235\\
yifan.chen@vanderbilt.edu
\And
Yevgeniy Vorobeychik \\
Washington University in St. Louis \\
St. Louis, MO 63130 \\
eug.vorobey@gmail.com
}
\maketitle
\begin{abstract}
Despite the considerable success of convolutional neural networks in a broad array of domains, recent research has shown these to be vulnerable to small adversarial perturbations, commonly known as adversarial examples.
Moreover, such examples have shown to be remarkably portable, or transferable, from one model to another, enabling highly successful black-box attacks.
We explore this issue of transferability and robustness from two dimensions: first, considering the impact of conventional $l_p$ regularization as well as replacing the top layer with a linear support vector machine (SVM), and second, the value of combining regularized models into an ensemble.
We show that models trained with different regularizers present barriers to transferability, as does partial information about the models comprising the ensemble.
\end{abstract}

\section{Introduction}
 Convolutional Neural Networks (CNNs)  have a dominant performance in a collection of domains, of which computer vision has been perhaps the most successful application\cite{NIPS2012_4824}. However, recently people have shown that CNN models are susceptible to small adversarial perturbations that have come to be termed \emph{adversarial examples (AEs)} \cite{goodfellow2015}.


As significantly, many studies have demonstrated the extensive portability, or \emph{transferability} of AEs across deep learning models \cite{DBLP:journals/corr/SzegedyZSBEGF13,DBLP:journals/corr/LiuCLS16}. More precisely, an AE generated against one model has been shown to commonly fool another deep learning model with similar architecture and trained independently either on a similar dataset, or using queries from the original model.
The phenomenon of transferability has been critical to the successful design of black-box attacks on deep learning, where the attacker need not have any knowledge of the model they are attacking to succeed \cite{DBLP:journals/corr/LiuCLS16,DBLP:journals/corr/PapernotMGJCS16}.


In our paper, we provide an in-depth study on how the hyperparameters of CNN models influenced the transferability of AEs generated from them. More importantly, we studied how transferability of AEs is influenced by the characteristic of target models and test models we chose. We summarize our primary contributions as the following: 
\begin{enumerate}
    \item Transferability of AEs generated from a target CNN model is low on other CNN models with different hyperparameters from the target CNN model. In particular, we show that conventional $l_p$ norm regularization can play a crucial role.
    \item AEs generated from a target ensemble model also have higher transferability on test ensemble models, if the target and test models shared common parts of CNN models.
    \item Transferability of AEs generated from a target \emph{ensemble} model is positively related to the number of \emph{sub-models} (see Table $1$ for definition) within the ensemble model. Moreover, such AEs tend to have greater transferability on CNN models compared to the amplified AEs generated from a CNN model, when the two kinds of AEs have the same noise magnitude (See table $1$ for definition).
\end{enumerate}

In the next section, we give a brief overview of related work to our discussion on transferability. We then discuss the methodology we used during our experiments. After that, we discussed the experiment setting and results. Finally, we provide our conclusion on transferability. We also provide all terminologies we used throughout the paper in Table~\ref{T:terminology}.

\begin{table}[h!]
\begin{tabular}{ |p{2cm}||p{5cm}|}
  \hline
Terminology&Definition\\
\hline
Target Model&Models whose architectures and gradient functions are used by attack algorithms to generate AEs.\\
\hline
Transferability&How well do the AEs generated from one or more target CNN models also trick non-target CNN models.\\
\hline
Ensemble Model& A model that is composed of one or more CNN models or variation of CNN models. The architecture is shown in figure $1$\\
\hline
Sub-model&A CNN or variation of CNN model within an ensemble model.\\
\hline
ASR (Attack Success Rate)&A measure for transferability of AEs generated from a single model. It is calculated using equation $(1)$ in this paper. Details are described in the methodology part.\\
\hline
AASR (Average Attack Success Rate) &A measure for the transferability of AEs generated from a group of models. It is calculated using equation $(2)$ in this paper. Details are described in the methodology part.\\
\hline
Noise Magnitude&A measure for the difference between a group of source images and the corresponding AEs. It is calculated using equation $(3)$ in this paper. Details are described in the methodology part.\\
\hline
\end{tabular}
\caption{Terminology}
\label{T:terminology}
\end{table}

\section{Related work}

In response to various attack methods, researchers have proposed many defensive methods against AEs targeting at neural network models \cite{DBLP:journals/corr/GuR14,DBLP:journals/corr/LuoBRPZ15,DBLP:journals/corr/PapernotMGJCS16,pmlr-v70-cisse17a}. However, our goal is to analyze how transferability is influenced by the nature of normal CNNs. Thus in our experiments, we study the transferability of AEs on models that are not augmented with any defensive methods. 

The work of \cite {DBLP:journals/corr/SzegedyZSBEGF13} show the low transferability of AEs generated from a target CNN model on other non-target CNN models that have different architectures. Based on their researches, we further find the low transferability of AEs generated from target CNN models on non-target models with different hyperparameters from the target models, while the architectures of the target CNN models and non-target CNN models are the same. 

In the work of \cite {DBLP:journals/corr/SzegedyZSBEGF13}, the author also suggests that more significant differences between source images and AEs can result in higher transferability. In our experiments, we show that by generating AEs from ensemble models, we can also increase the transferability without significantly increasing the differences between source images and AEs.

Based on researches on transferability, \cite{DBLP:journals/corr/LiuCLS16} manages to develop a new black-box attacks methods which are based on ensemble models to create AEs with higher transferability. Our research instead analyzes how to generate AEs with high or low transferability using existing attack methods by manipulating target models.

\section{Methodology}
In this section, we describe how we generate models for our experiments, how we evaluate the results in multiple measures, and how we design our experiments.
\subsection{Model Generation}
The following sub-subsections describe how we generated different types of models. We used the CIFAR-10 dataset \cite{krizhevsky2009learning} for training models.
\subsubsection{ResNet-20}
We use the ResNet architecture to described by \cite{DBLP:journals/corr/HeZRS15} to generate our own ResNet-20.

\subsubsection{ResNet-SVM-20}
Using the technique mentioned in \cite{DBLP:journals/corr/Tang13}, we can generate the model ResNet-SVM-20. The model is just a ResNet-20 model with an SVM layer generated from the ResNet-20. We retrain the model on the CIFAR-10 dataset.

\subsubsection{Convolutional Neural Networks with Regularization}
We can also generate regularized ResNet-20 and regularized ResNet-SVM-20 models by adding $L1$ and $L2$ regularization to the ResNet-20 and ResNet-SVM-20 \cite{regularization}. The regularized models are retrained on the CIFAR-10 dataset. The naming of a regularized model is the base model name, ResNet-20 or ResNet-SVM-20, followed by the regularization types, $L1$ or $L2$. For example, ResNet-20-L1 stands for a ResNet-20 model with $L1$ regularization.

\subsubsection{Ensemble Model Generation}
The architecture of an ensemble model is shown in figure $1$. An ensemble model takes images as inputs and feeds them into each sub-model within it. Then it takes the average of sub-models' weighted classification results as its final classification results for images. We use the following criteria in our experiments to generate ensemble models:
\begin{itemize}
    \item Each sub-model within an ensemble model is a ResNet-20 or ResNet-20-SVM model, with or without regularization.
    \item An ensemble model can be composed of only one sub-model.
    \item We do not retrain ensemble models since the accuracy of each ensemble model is already close to optimal (i.e., larger than that of each its sub-models). 
    \item We use $1$ for all weights throughout the paper for simplicity of experiments and discussion.
\end{itemize}  

 \begin{figure}[htbp]
  \centering
  \includegraphics[scale=0.5]{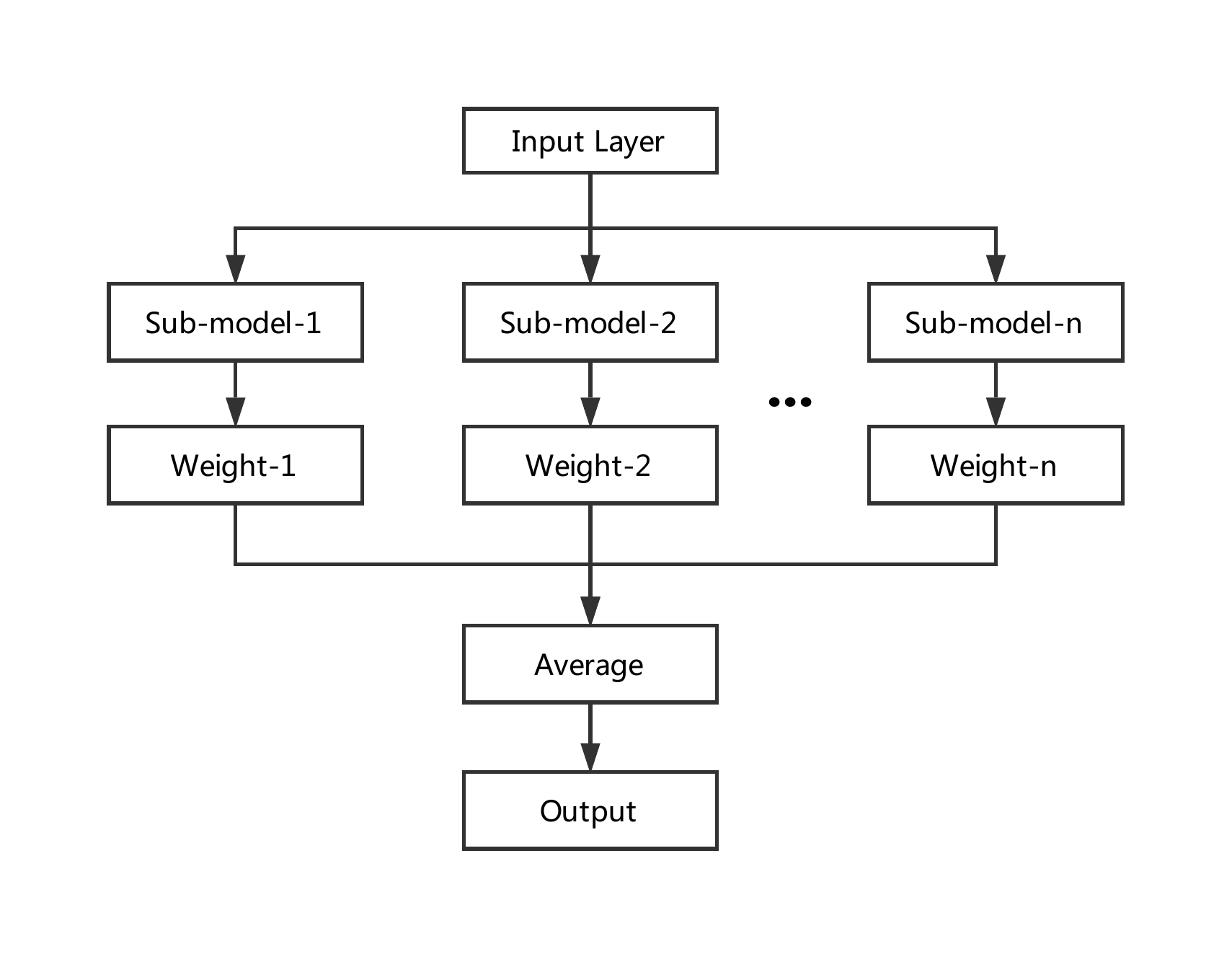}
  \caption{The architecture for an ensemble model we used in our experiments.}
  \label{fig:Ensemble_Model.png}
   \end{figure}

\subsubsection{Generating Adversarial Examples}

In our experiments, we tried several methods to generate AEs, including FGSM (fast gradient sign method) \cite{goodfellow2015}, FGM (fast gradient method) \cite{goodfellow2015} and DeepFool using $l_2$ norm \cite{DBLP:journals/corr/Moosavi-Dezfooli15}. We tried all three algorithms to generate one AE for each test image from CIFAR-10. Due to limited space and similar result patterns for three methods, in the later section, we only present the results for DeepFool using $l_2$ norm. \\
All methods mentioned above require the gradient function of the target model to generate AEs against the target model. When we choose an ensemble model as the target model, the gradient function used by attack algorithms is just the sum of the gradient function of each sub-models within the ensemble model. \\
If a source image is already wrongly classified by the target model, it is regarded as an AE for the target model and thus not modified. \\

 \begin{figure*}[h!]
  \centering
  \includegraphics[scale=0.65]{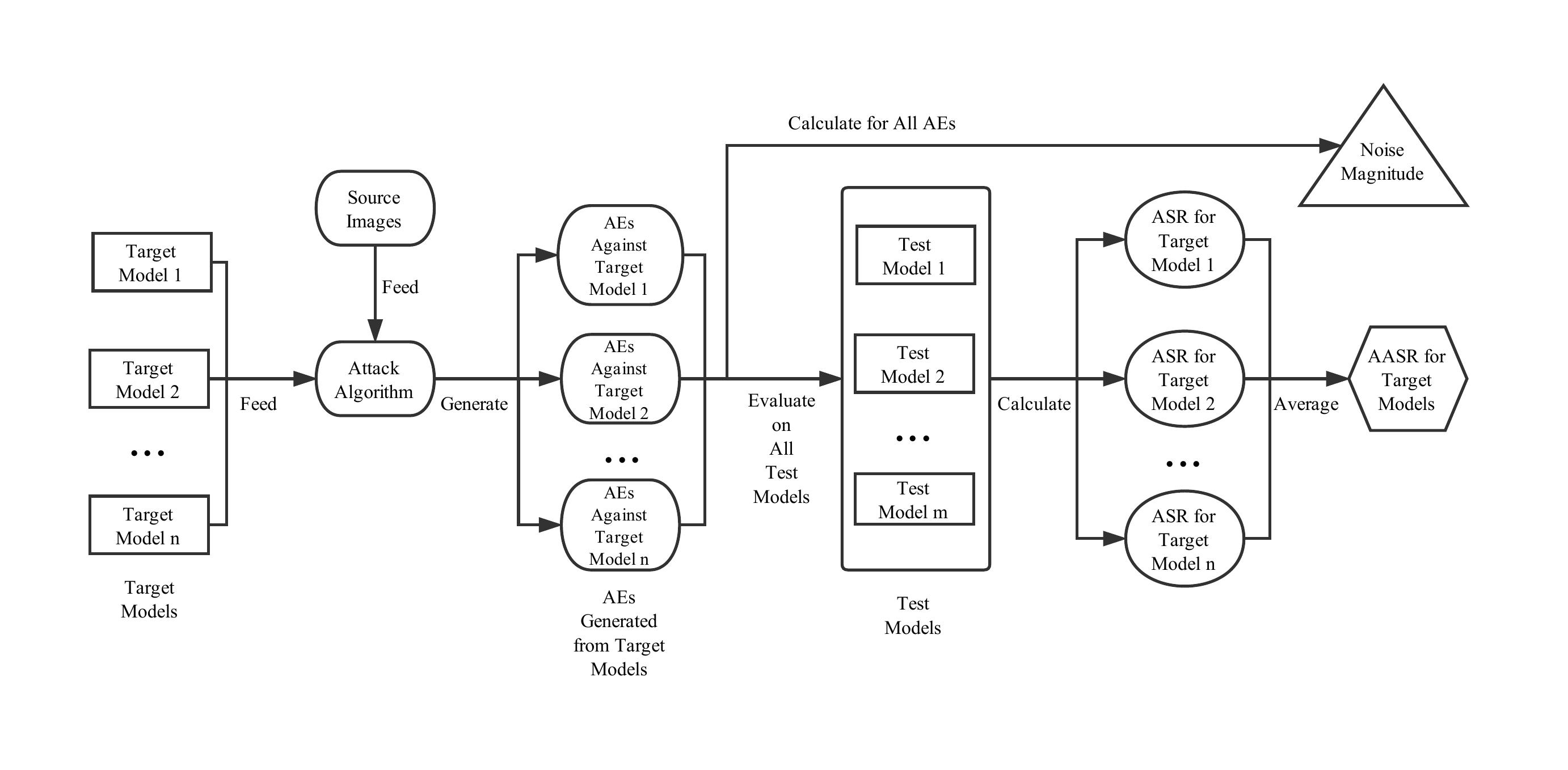}
  \caption{This is the experiments procedure we follow for all experiments we have done.}
  \label{fig:Experiment.pdf}
   \end{figure*}

\subsection{Evaluation Metrics}
In this subsection, we describe how we evaluate the transferability and noise magnitude of AEs in our experiments.

\subsubsection{Measure of Transferability}
In our experiments, the transferability of AEs generated from a target model on test models is measured by its \emph{ASR} (Attack Success Rate) on a set of test models, which should all be different from the target models. The ASR for AEs means the portion of AEs that successfully trick test models. The larger ASR of AEs on the test models indicates the larger transferability of AEs on the set of test models. To calculate ASR, we define the following variables:
\begin{itemize}
    \item $x$: We label each test model with an unique integer $x$, where $x \in \{1,2,3, ..., m\}$. $m$ is the total number of test models.
    \item $a_x$: $a_x$ stands for the number of AEs that were generated from the target model and successfully tricked the test model $x$,
    \item $A$: $A$ stands for the total number of AEs generated from the target model. In our case, this is always $10000$, which is the number of test images in the CIFAR-10 dataset.
    \item $S$: $S$ stands for the ASR of AEs evaluated on the set of test models.
\end{itemize}
Then, we calculate the ASR of AEs generated from a target model on $m$ test models as follows:
  \begin{equation}
  S = \frac{(a_1 / A + a_2 / A + ... a_x / A)}{m}
  \end{equation}

In our experiments, we mostly focus on the transferability of AEs generated from a group of target models. The groups are formed in such a way that target models within a group share one group property. Evaluating AEs generated from such group of models together help us to find the relationship between transferability and the group property. Therefore we use \emph{AASR} (Average Attack Success Rate), denoted as $\mu$, to evaluate the transferability of AEs generated from a group of target models. The AASR is just the average of ASRs for multiple models.

To calculate the AASR of $n$ target models, we first use a unique integer $y$, where $y \in \{1, 2, 3, ..., n\}$, to label each target model within the group. The ASR for AEs generated from the target model $y$ is denoted as $S_y$. $S_y$ can be calculated using equation $(1)$. The AASR from a group of $n$ target models is calculated as follows:
  \begin{equation}
  \mu = \frac{(S_1 + S_2 + ... + S_n)}{n}
  \end{equation}
The higher AASR indicates higher transferability of AEs generated from a target model group on the set of test models. For simplicity of discussion, we use AASR for evaluating AEs generated from a target model, which has a group size of $1$.

\subsubsection{Measure of Noise Magnitude}
The transferability of AEs are also influenced by the magnitude of the noise added to the source images to generate AEs according to \cite{DBLP:journals/corr/SzegedyZSBEGF13}. However, multiple measures have been used to describe the \emph{noise magnitude} \cite{DBLP:journals/corr/SzegedyZSBEGF13,DBLP:journals/corr/LiuCLS16}. \\
In our paper, we define the noise magnitude, denoted as $M$, for AEs to be the average sum of absolute difference between each adversarial image and the corresponding source image per pixel per image. We then define variables for our equation as follows:
\begin{itemize}
    \item $o$: $o$ is the number of source images and also the number of AEs generated from the them.
    \item $p_i$: A source image labeled by $i$, where $i \in \{1, 2, 3, ..., o\}$
    \item $q_i$: An AE generated from $p_i$, where $i \in \{1, 2, 3, ..., o\}$
    \item Dimension: The image dimension for each source image or AE is $c * r * 3$.
\end{itemize}

Therefore, the noise magnitude for $o$ source images and the corresponding AEs is given by:
  \begin{equation}
  M = \frac{\sum_{k=1}^{c}\sum_{j=1}^{r}{\sum_{i=1}^{o}{|p_i - q_i|)}}}{o * c * r}
  \end{equation}
The larger noise magnitude of AEs indicate that the AEs are more different from the source images.

\subsection{Experiment Design}
All our experiments contain three uniform steps and one optional step:
\begin{enumerate}
    \item \textbf{Generate Target Models and AEs}: We choose target models to generate AEs from them. We then generate AEs from each model. Target models and the corresponding AEs may be divided into different groups based on some properties of target models.
    \item \textbf{Generate Test Models}: For each target models, we specify the corresponding test models. Test models may also be divided into groups.
    \item \textbf{Evaluation on Transferability}: We calculate the AASR for AEs generated from each group of target models on each group of test models independently using equation $(2)$. Therefore if we have $3$ groups of target models and $4$ groups of test models, we can get $12$ AASR results in total.
    \item \textbf{Evaluation on Noise Magnitude}: We also evaluate the noise magnitude of AEs using equation $(3)$ in some experiments.
\end{enumerate}
We summarize these steps in figure $2$. In the next section, we only present the first two steps for each experiment. The third step and fourth step are the same for all experiments and thus omitted. The results are presented in figures.

\section{Experiment Result and Discussion}
In the next three subsections, we present our experiments and discussion about how the transferability of AEs changed in various conditions. All experiments follow the procedure described in Experiment Design and figure $2$. Before walking through the experiments and discussions, We first introduce some \emph{common models} that we used for all experiments. 

We trained ResNet-20 on the CIFAR-10 dataset to reach an accuracy of $90.57\%$. The ResNet-SVM-20 model trained on the CIFAR-10 dataset reach an accuracy of $90.82\%$. We directly refer to them in later experiments.

Then, we generated $20$ regularized models with various regularization parameters, which included $5$ $L1$ and $5$ $L2$ regularized models generated from ResNet-20 and ResNet-SVM-20 respectively. The $L1$ regularization parameters were chosen from $[0, 5]$ in equal logarithmic space. The $L2$ regularization parameters were chosen from $[0, 35]$ in equal logarithmic space. Each regularized models had an accuracy higher than $89\%$ over the CIFAR-10 dataset. We classified all regularized models into four groups based on the final layer and regularization types: \emph{ResNet-L1}, \emph{ResNet-L2}, \emph{ResNet-SVM-L1}, \emph{ResNet-SVM-L2}. We refer to these regularized models by their group names in later experiments.

\subsection{Influence of Hyperparameters on Transferability}
In this subsection, we evaluate how the transferability of AEs generated from CNN models is influenced by hyperparameters, including the final layer types, the regularization types, and the regularization parameters. Our results indicate that AEs generated from a target model have low transferability on models with the same architectures but different hyperparameters from it.

\subsubsection{Experiment 1: Various Final Layer Types} 
To evaluate the influence of the final layer types on transferability of AEs, we set up our experiment $1$ as follows:
\begin{enumerate}
    \item \textbf{Target Models}: We selected ResNet-20 and ResNet-SVM-20 from the common models. We also selected ResNet-23, ResNet-SVM-23, ResNet-26 and ResNet-SVM-26 generated with the same procedure mentioned in methodology part. AEs were generated from each model.
    \item \textbf{Test Models}: For each target model, the test models were itself and the model with the same layer number but different final layer types from it. For example, test models for ResNet-20 were ResNet-20 and ResNet-SVM-20.
\end{enumerate}
We show the AASR for AE generated from models with different final layers in figure $3$. 

 \begin{figure}[h!]
  \centering
  \includegraphics[scale=0.6]{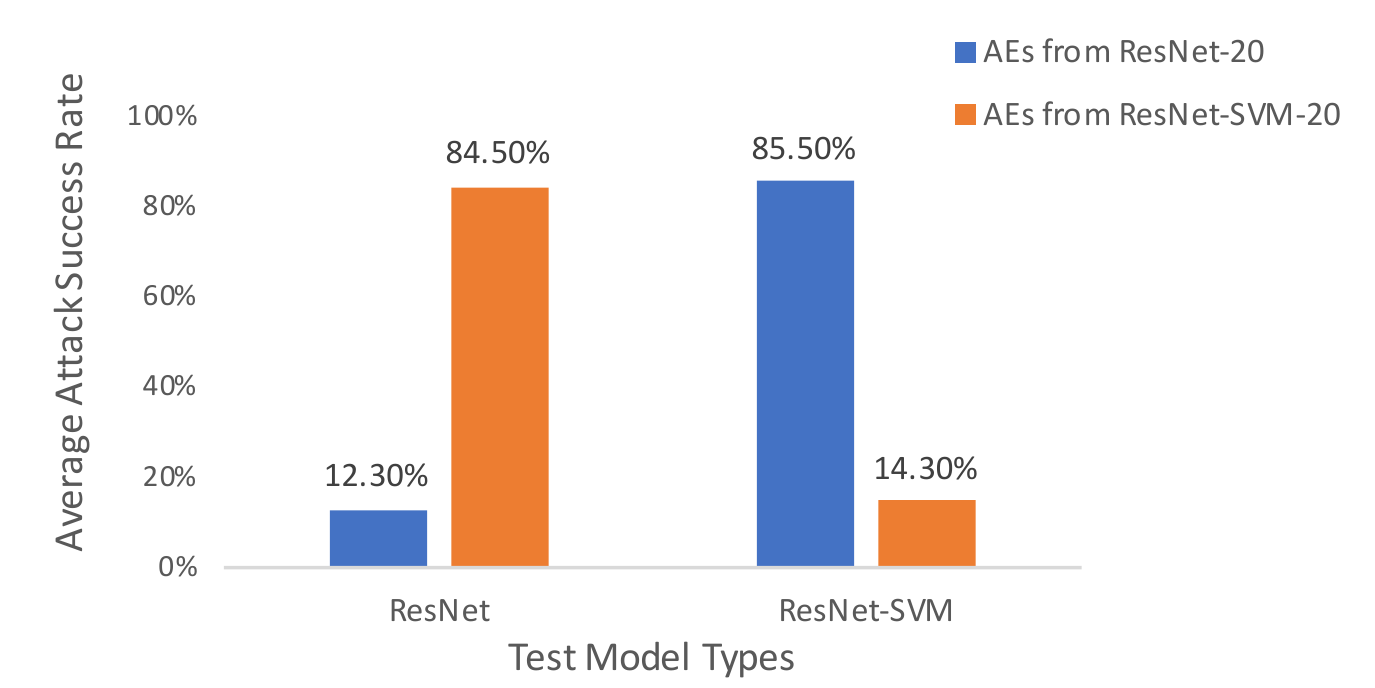}
  \caption{The AASR of AEs generated from ResNet-20 and ResNet-SVM-20.}
  \label{fig:Transferability_VS_Activation_Types.pdf}
   \end{figure}

\subsubsection{Experiments 2: Various Regularization Types} To evaluate how the transferability of AEs varies with multiple regularization types, we set up our experiment $2$ as follows: 
\begin{enumerate}
    \item \textbf{Target Models}: We selected all four groups of regularized ResNet and ResNet-SVM models from common models. AEs were generated from each model and also divided into four groups based on the model that they were generated from.
    \item \textbf{Test Models}: For each group of target models, the test models were exactly the four groups of regularized models selected in target models. 
\end{enumerate}
The results of AASR for AEs generated from each group of target models are shown in figure $4$.

\begin{figure*}[h!]
  \centering
  \includegraphics[scale=0.6]{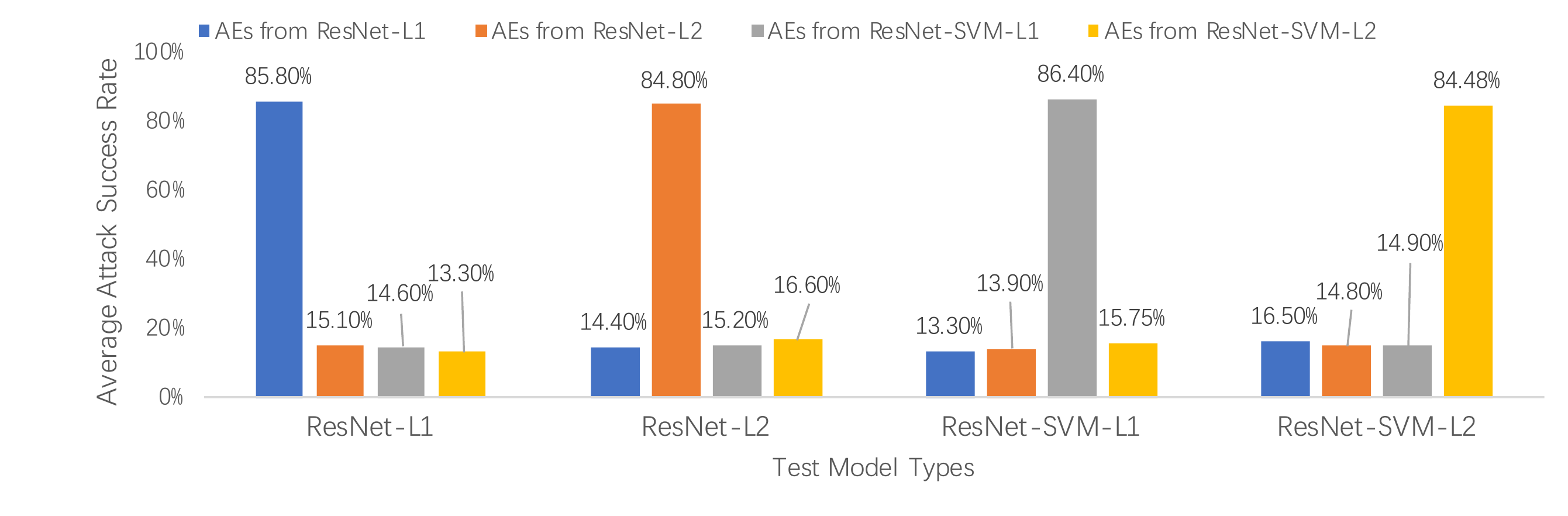}
  \caption{The AASR of AEs generated from ResNet-20 and ResNet-SVM-20 models with various regularization types.}
  \label{fig:Transferability_VS_Regularization_Types.pdf}
   \end{figure*}

\subsubsection{Experiments 3: Various Regularization Parameters} To evaluate how the transferability of AEs varies with multiple regularization types, we set up our experiment $3$ as follows: 
\begin{enumerate}
    \item \textbf{Target Models}: Three ResNet-L2 models with regularization parameters $5, 4.999999, 4.9999999$.
    \item \textbf{Test Models}: For each target model, the test models were the all models in target models.
\end{enumerate}
The results of AASR for AEs generated from each group of models are shown in figure $5$.

\begin{figure}[htbp]
  \centering
  \includegraphics[scale=0.45]{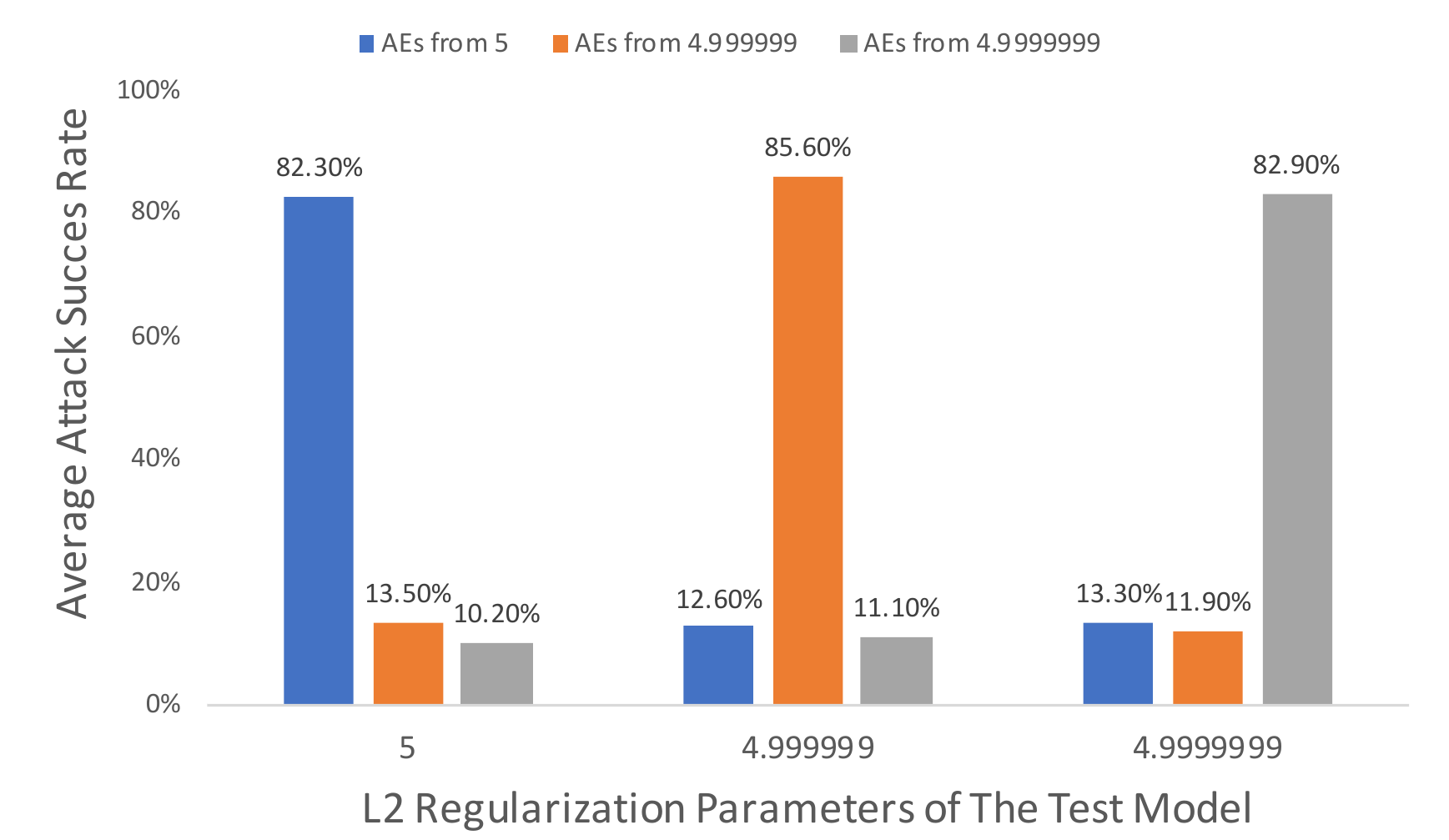}
  \caption{The AASR of AEs generated from ResNet-20 models with various regularization parameters.}
  \label{fig:Transferability_VS_Regularization_Parameters.pdf}
   \end{figure}

\subsubsection{Discussion} From the three figures, we can observe that the AASR of AEs generated from ResNet models is low on other ResNet models with different final layer types, regularization types, and parameters from them. This phenomenon indicates that the transferability of AEs generated from target models is low on models that have different hyperparameters from the target models. We believe CNN models with other architectures should follow the patterns similar to what we observed here since no unique properties of ResNet-20 were involved in the experiments.

To account for such low transferability, we hypothesize that gradient-based attacks tend to utilize the unique decision features of each model to produce AEs with minimum noise. However, the retraining procedure after we change hyperparameters cause modified models to have different decision features from the original. The new decision features thus make the modified models not vulnerable to the most AEs generated from the original model. 

\subsection{Transferability of Adversarial Examples Generated From Ensemble Models, Part 1}
In this subsection, we evaluate how the transferability of AEs on ensemble models is influenced by two factors, including shared sub-models between ensemble models and redundant sub-models. The shared sub-models here means a target ensemble model and a test ensemble model have sub-models that are identical. Redundant sub-models means that a target ensemble model has sub-models that are not part of the test model, besides having all sub-models of the test model.

\subsubsection{Experiments 4: Shared Sub-models between Ensemble Models} 
To evaluate how shared sub-models influence transferability, we set up our experiment $4$ as follows: 
\begin{enumerate}
    \item \textbf{Target Models}: We selected $3$ models from $5$ ResNet-L1 and $3$ models from $5$ ResNet-L2. The $4$ other models were not selected. Then we generated all possible ensemble models from the $6$ selected ResNet models. Each ensemble model had $x$ sub-models inside it, where $x \in [1, 6]$. The target models included all ensemble models and the $4$ models which were not selected.
    \item \textbf{Test Models}: The test model was the largest ensemble model, which was composed of all $6$ models we picked.
\end{enumerate}
The corresponding AASR results are shown in figure $6$. It is worth noting that the $4$ regularized ResNet models not used to generate ensemble models shared $0$ identical sub-models with the test ensemble model.

\begin{figure}[htbp]
  \centering
  \includegraphics[scale=0.6]{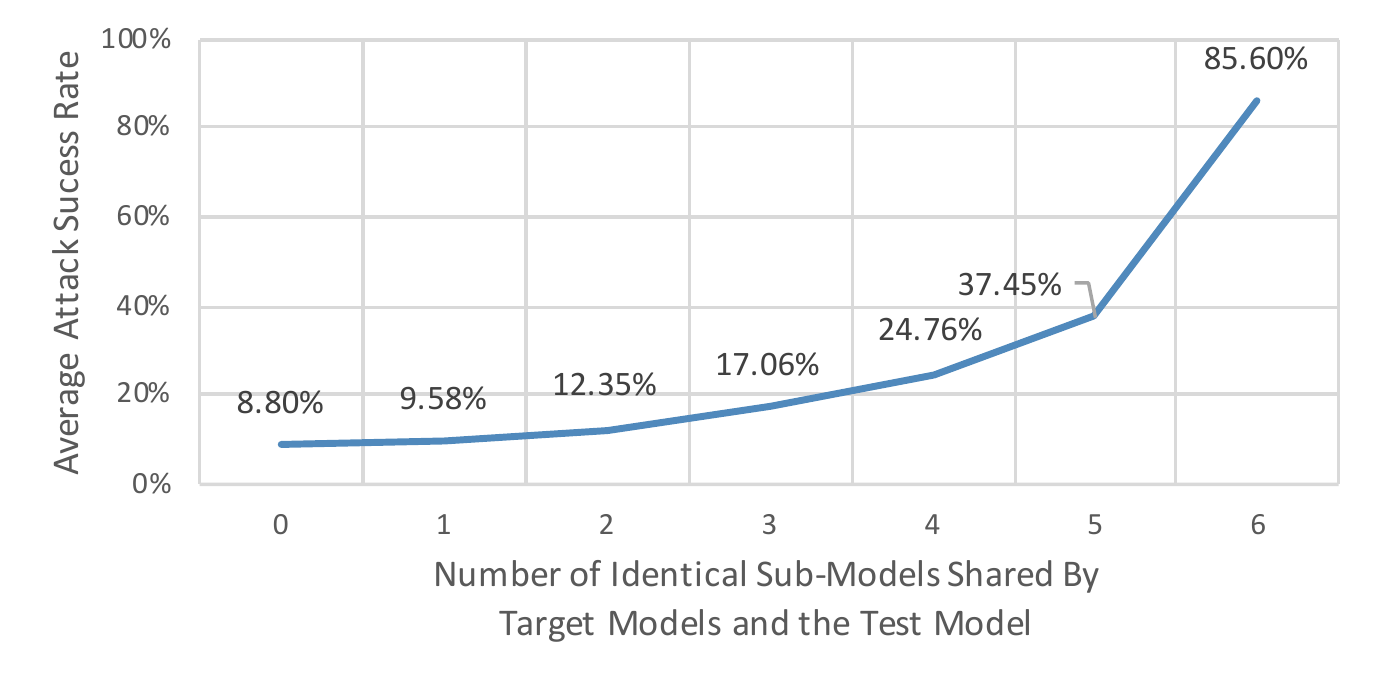}
  \caption{The AASR of AEs generated from ensemble models. The results are evaluated on the ensemble model with $6$ sub-models inside it.}
  \label{fig:Accuracy_VS_Attacker_knowledge.pdf}
   \end{figure}
   
\subsubsection{Experiments 5: Redundant Sub-Models}
To evaluate how redundant sub-models influence transferability, we set up our experiment $5$ as follows: 
\begin{enumerate}
    \item \textbf{Target Models}: For each test model, we selected all target models from the experiment $4$'s target ensemble models such that each target model contained the test model as its sub-model. 
    \item \textbf{Test Models}: A regularized ResNet model that was used to build ensemble models in the experiment $4$. 
\end{enumerate}
We repeated the procedure for multiple test models, which included all regularized ResNet models not selected in the experiment $4$. The corresponding AASR results are shown in figure $7$. 

 \begin{figure}[htbp]
  \centering
  \includegraphics[scale=0.6]{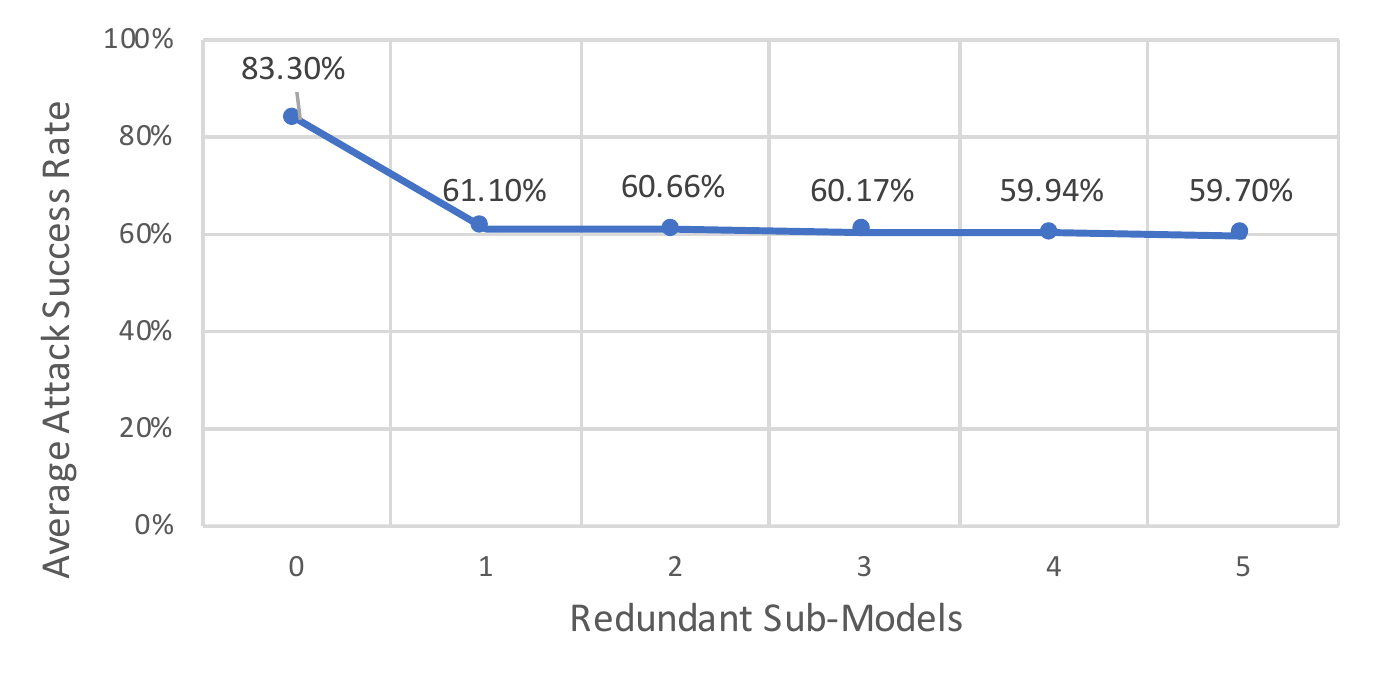}
  \caption{The AASR of AEs generated from ensemble models. The results were evaluated on $6$ regularized ResNet-20 models}
  \label{figRedundant.pdf}
  \end{figure}

\subsubsection{Discussion} In the experiment $4$, the increasing AASR indicates that the transferability of AEs generated from a target model grows with more sub-models shared by the target model and the test ensemble models. Also, the growth of transferability becomes more substantial with more identical sub-models shared by the target model and the test ensemble models. In the experiment $5$, the result indicates that redundant sub-models within an ensemble model neither decreases nor increases the transferability. 

\subsection{Transferability of Adversarial Examples Generated From Ensemble Models, Part 2}
In this subsection, we evaluate how the number of sub-models within an ensemble model influences the transferability of AEs generated from it on ResNet models. 

\subsubsection{Experiments 6: Transferability Evaluated on Regularized ResNet models} 
We set up our experiment as follows:
\begin{enumerate}
    \item \textbf{Target Models}: We selected $8$ models, $4$ from ResNet-L1 and $4$ from ResNet-L2, to generate all possible unique ensemble models. The $2$ other models from the two groups of models were not selected. Each ensemble model was composed of $x$ sub-models, where $x \in [1, 8]$. We then divided ensemble models selected into $8$ groups by the number of sub-models within each model.
    \item \textbf{Test Models}: The other $2$ regularized ResNet-20 models from ResNet-L1 and ResNet-L2 were used as the test models.
\end{enumerate}
We show the AASR of AEs evaluated on the $2$ test models in figure $8$.
\begin{figure}[htbp]
  \centering
  \includegraphics[scale=0.6]{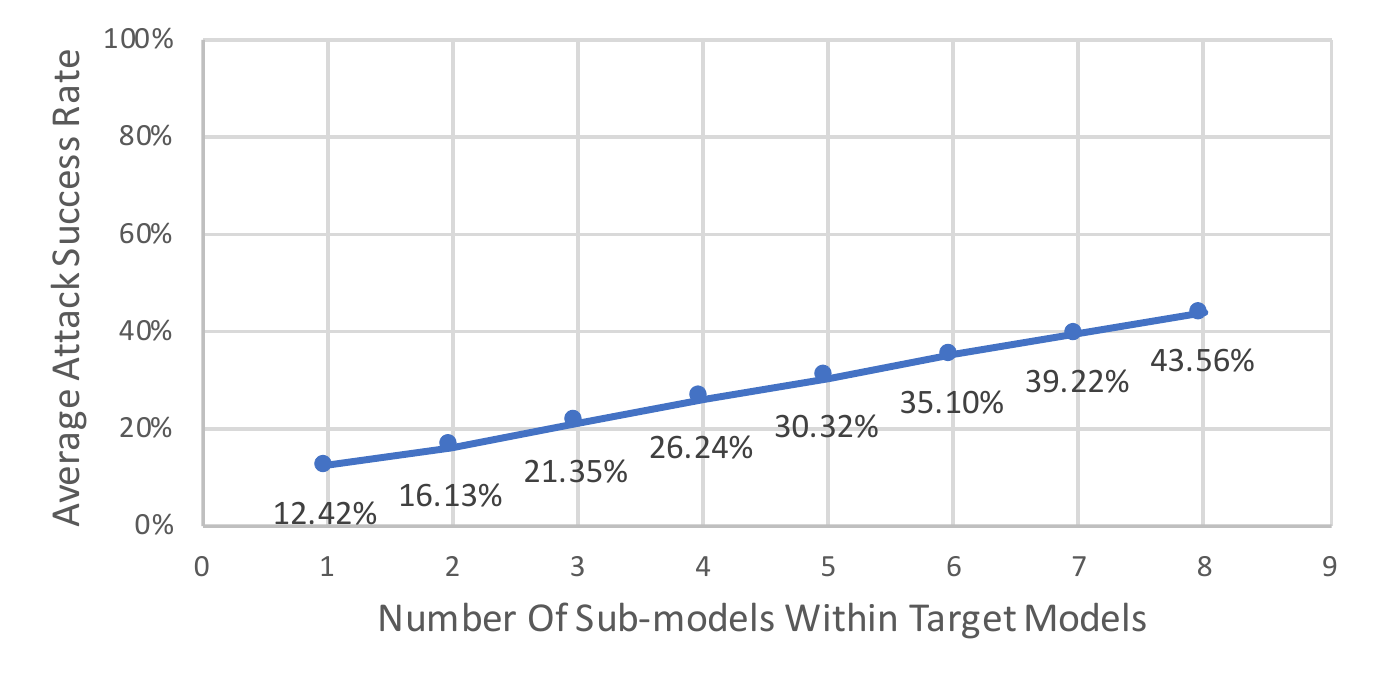}
  \caption{The AASR of AEs generated from ensemble models with various numbers of sub-models. The results were evaluated on regularized ResNet-20 models which were not used to build the target ensemble models.}
  \label{fig:Transferability_VS_Sub_Model_Number.pdf}
   \end{figure}

From the figure, we can observe that the AASR of AEs on regularized ResNet-20 models increases with respect to the number of sub-models inside an ensemble model. However, from the previous section, we know that transferability generated from a target model is supposed to be low on other models with different regularization types and parameters from the target model. In other words, by generating AEs from ensemble models, we can generate AEs with larger transferability on regularized models without modification to the attack algorithms.

\subsubsection{Experiment 7: Comparison with Amplified Adversarial Examples Generated from ResNet-20 Models}
However, as mentioned in \cite{DBLP:journals/corr/SzegedyZSBEGF13}, the magnitude of noise added to the sources images also influence the transferability of AEs. The noise magnitudes for different groups of AEs generated in experiment $6$ is shown in figure $9$.

\begin{figure}[h!]
  \centering
  \includegraphics[scale=0.6]{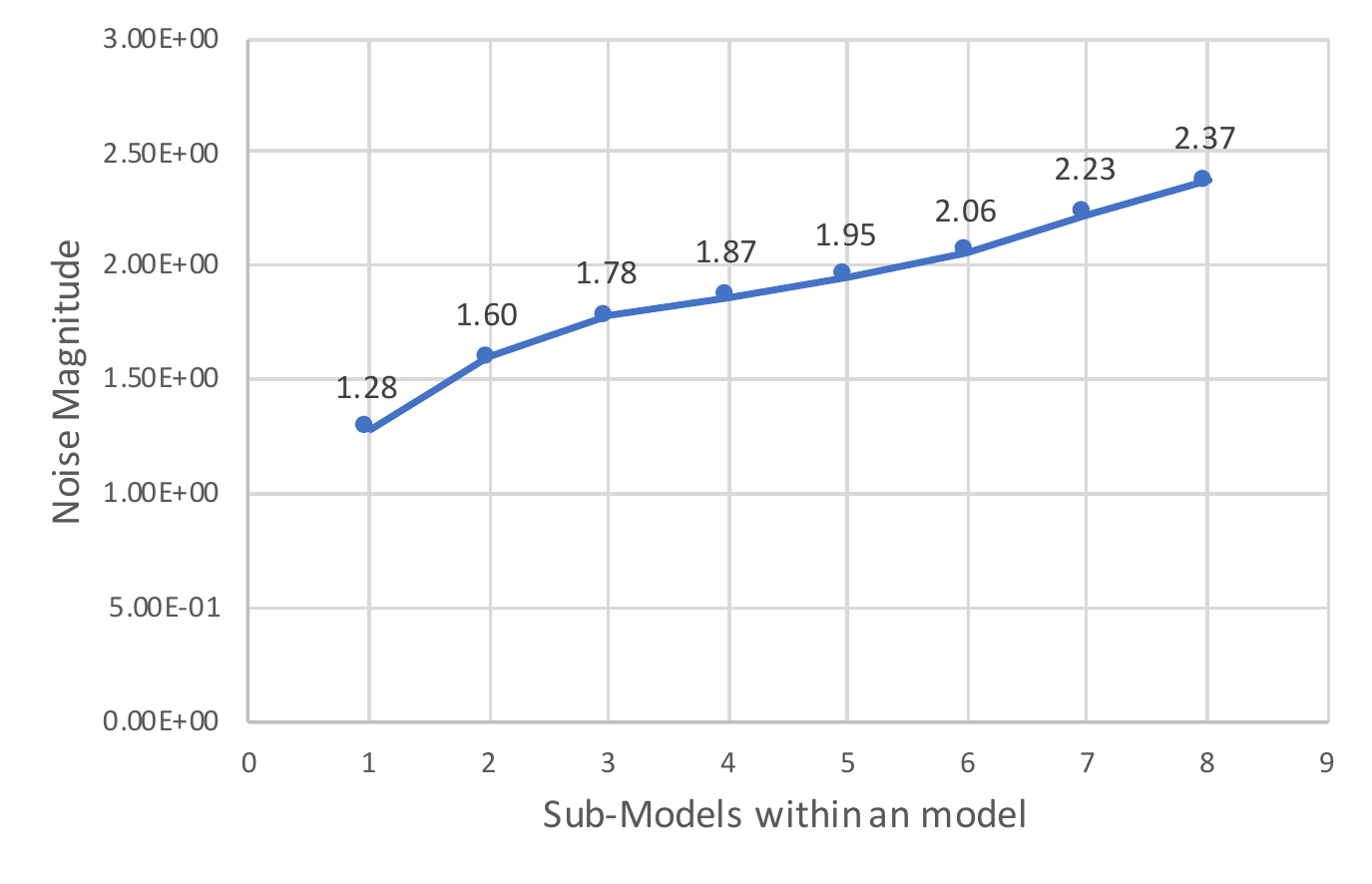}
  \caption{The average noise magnitude for each group of AEs evaluated in figure $8$.}
  \label{fig:Noise_Magnitude.pdf}
   \end{figure}

To eliminate the different noise magnitudes, we manually amplified AEs generated from each model to have the same magnitude. We set up our experiment as follows:
\begin{enumerate}
    \item \textbf{Target Models}: The target models used in experiment $6$. The noise magnitude of all AEs generated from target models was increased to $2.37$, which was the noise magnitude of the ensemble model with $8$ sub-models in it. 
    \item \textbf{Test Models}: The test models used in experiment $6$.
\end{enumerate}
The result of AASR is shown in figure $10$. The result showed that the more sub-models within an ensemble model gave AEs generated from it larger transferability compared to amplified AEs generated from smaller ensemble models, even when they have the same noise magnitude.

\begin{figure}[h!]
  \centering
  \includegraphics[scale=0.6]{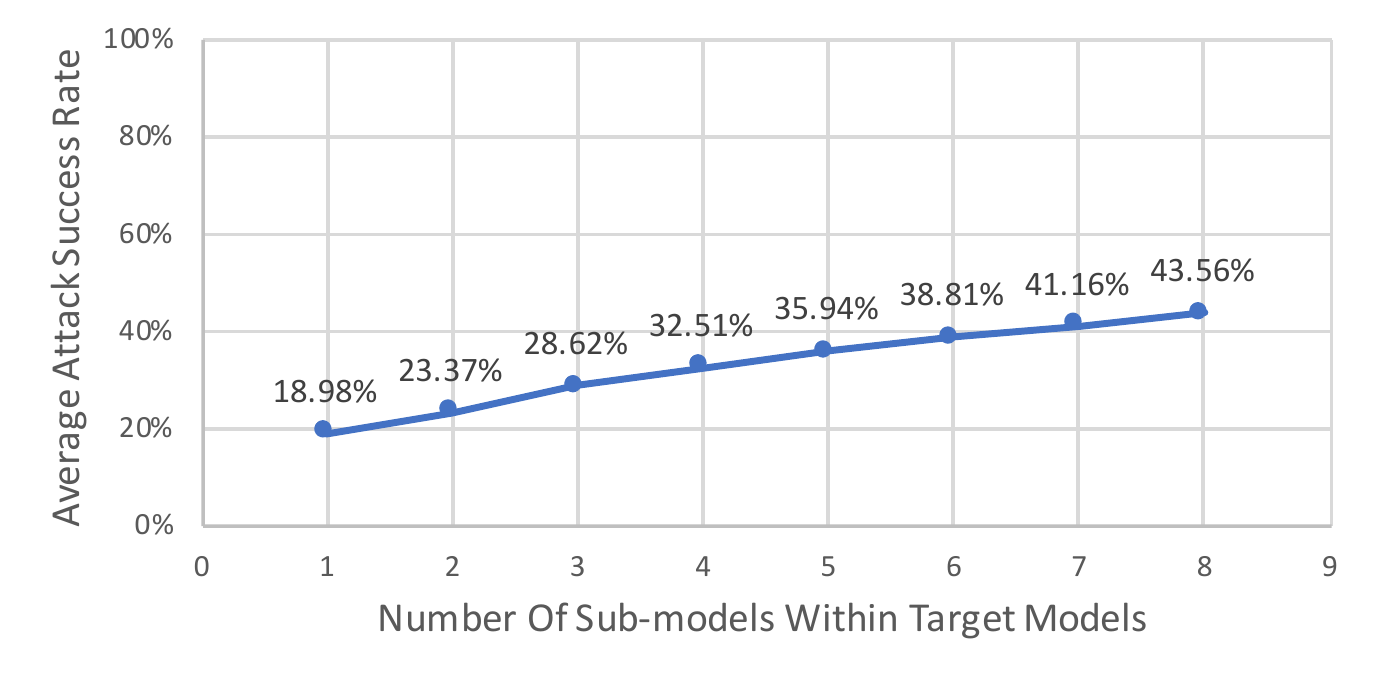}
  \caption{The AASR of amplified AEs generated from ensemble models with various numbers of sub-models. The results were evaluated on regularized ResNet-20 models which were not used to build the target ensemble models.}
  \label{figTransferability_the_same_magnitude.pdf}
   \end{figure}
   
\begin{figure*}[ht!]
  \centering
  \includegraphics[scale=0.6]{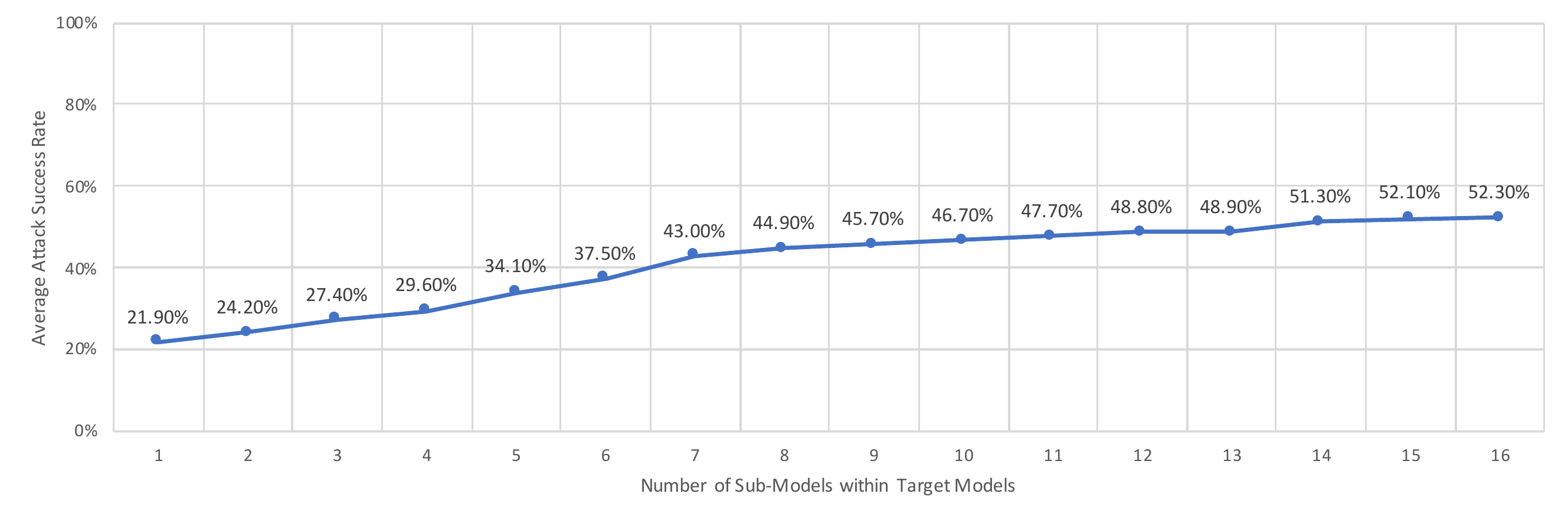}
  \caption{The AASR of AEs generated from ensemble models with various numbers of regularized ResNet-SVM models. The results were evaluated on regularized ResNet-20 models which were not used to build the target ensemble models.}
  \label{fig:limit.pdf}
   \end{figure*}
   
\subsubsection{Experiement 8: The Limit of Transferability Growth}
To test if the transferability of AEs grows unlimited with more sub-models inside the target ensemble models, we set up our experiment as follows:
\begin{enumerate}
    \item \textbf{Target Models}: We generated $18$ regularized ResNet-20 models, where $9$ of them were L1 regularized, and $9$ of them were L2 regularized. We then generated ensemble models that were composed of $x$ sub-models, where $x \in [1, 18]$. Due to computation limitations, we only generated $3$ ensemble models for each $x$. Again, we divided them into $18$ groups based on the number of sub-models within each ensemble model.
    \item \textbf{Test Models}: We chose the $1$ regularized ResNet-20 models from ResNet-L2 which have different regularization parameters from any model selected to generate the target ensemble models.
\end{enumerate}

We repeated the experiment several times with various target ensemble models and test models with the same criteria mentioned in the experiment settings. Only one AASR result is shown in figure $11$ due to space limitations. From the results, we can only conclude that the transferability of AEs grows on regularized ResNet models with increasing sub-models inside an ensemble model. The growth speed slows down as the number of sub-models increases.

\subsubsection{Discussion} The larger AASR indicates that AEs generated from larger ensemble models still have higher transferability compared to amplified AEs generated from smaller ensemble models, even when they have the same magnitude of noise. 

To account for the higher transferability of AEs generated from large models on CNN models, we hypotheses that the sum of gradients from each sub-models may have canceled out some unique features within each gradient function of sub-models. The more sub-models in the ensemble models, the fewer unique features in the sum of the gradient function. Thus, AEs generated from ensemble models utilize more general features from gradient functions and thus have higher transferability on regularized CNN models.

We also learn that increasing AEs' transferability becomes computationally more expensive as the transferability of it grows. Also, Amplifying existing AEs takes significantly less computation than generating AEs from large ensemble models. Thus, we may want to amplify the AEs generated from small ensemble models to balance between minimal perturbation and computation efficiency. 

\section{Conclusion}
In our paper, we provide an in-depth study on the transferability of AEs generated from different kinds of models, especially those generated from ensemble models. Our results indicate that AEs from a CNN model are largely not transferable to other CNN models with various hyperparameters. Additionally, the results indicate that AEs generated from a target ensemble model tend to have higher transferability on CNN models when the target model is composed of more sub-models. Finally, we also find that more identical sub-models shared between target and test ensemble models could help increase the transferability of AEs generated from the target model. We also include our hypothesis for these phenomena. To summarize, our results identify the various factors that can influence the transferability of AEs generated from CNN models and ensemble models. Our findings on transferability can also help generate adversarial examples with the desired transferability on neural networks.

\section{Acknowledgement}
This work was supported by the Vanderbilt Undergraduate Summer Research Program. We also would like to thank Dr.Fermanda Eliott for revising this paper.

\bibliography{cite.bib}
\bibliographystyle{aaai}

\end{document}